\pdfoutput=1

\documentclass[11pt]{article}

\usepackage[final]{acl}

\usepackage{times}
\usepackage{latexsym}

\usepackage[T1]{fontenc}

\usepackage[utf8]{inputenc}

\usepackage{microtype}

\usepackage{inconsolata}
\usepackage{algorithm}
\usepackage{algorithmic}
\usepackage{subfigure}
\usepackage{makecell}

\usepackage{multirow}
\usepackage{makecell}
\usepackage{subfigure}
\usepackage{multicol}
\usepackage{multirow}
\usepackage{url}
\usepackage{amsmath,amsfonts,amsthm} 
\usepackage{graphicx}
\usepackage{subfigure}
 \usepackage{balance}
\usepackage{color}
\usepackage{tabularx}
\usepackage{array}
\usepackage{booktabs}
\usepackage{amssymb}
\usepackage{bbding}
\usepackage{bbm}
\usepackage[subfigure]{tocloft}
\title{Distillation Enhanced Generative Retrieval}

\author{Yongqi Li$^{1}$, Zhen Zhang$^{2}$, Wenjie Wang$^{2}$, Liqiang Nie$^{3}$, Wenjie Li$^{1}$,  Tat-Seng Chua$^{2}$ \\
        $^{1}$The Hong Kong Polytechnic University \\ 
        $^{2}$National University of Singapore
        $^{3}$Harbin Institute of Technology (Shenzhen)\\
        \texttt{\{liyongqi0,cristinzhang7,wenjiewang96,nieliqiang\}@gmail.com} \\ 
        \texttt{cswjli@comp.polyu.edu.hk dcscts@nus.edu.sg}}

\begin{document}
\maketitle
\begin{abstract}
Generative retrieval is a promising new paradigm in text retrieval that \textit{generates} identifier strings of relevant passages as the retrieval target. This paradigm leverages powerful generative language models, distinct from traditional sparse or dense retrieval methods. In this work, we identify a viable direction to further enhance generative retrieval via distillation and propose a feasible framework, named DGR. DGR utilizes sophisticated ranking models, such as the cross-encoder, in a teacher role to supply a passage rank list, which captures the varying relevance degrees of passages instead of binary hard labels; subsequently, DGR employs a specially designed \textit{distilled RankNet} loss to optimize the generative retrieval model, considering the passage rank order provided by the teacher model as labels. This framework only requires an additional distillation step to enhance current generative retrieval systems and does not add any burden to the inference stage. We conduct experiments on four public datasets, and the results indicate that DGR achieves state-of-the-art performance among the generative retrieval methods. Additionally, DGR demonstrates exceptional robustness and generalizability with various teacher models and distillation losses.
\end{abstract}

\section{Introduction}
Text retrieval is a crucial task in information retrieval and has a significant impact on various language systems, including search ranking~\cite{nogueira2019passage}, open-domain question answering~\cite{chen2017reading}, and retrieval augmented generation (RAG)~\cite{lewis2020retrieval}. In recent years, dense retrieval~\cite{lee2019latent,karpukhin2020dense} has been the dominant approach for text retrieval based on the advancements in encoder-based language models, like BERT~\cite{kenton2019bert}.  

With the advancement of generative large language models~\cite{brown2020language}, generative retrieval emerges as an alternative paradigm to dense retrieval. Generative retrieval leverages autoregressive language models to \textit{generate} identifier strings of target passages, such as Wikipedia page titles, to complete the retrieval process. Current approaches focus on exploring various identifiers to better represent passages. Initially, generative retrieval utilized page titles~\cite{de2020autoregressive} as identifiers but was limited to specific retrieval domains, such as Wikipedia. Subsequently, a range of identifier types were introduced, including numeric IDs~\cite{tay2022transformer}, substrings~\cite{bevilacqua2022autoregressive}, codebooks~\cite{sun2023learning}, and multiview identifiers~\cite{li2023multiview}, to consistently enhance generative retrieval for broader search scenarios and larger retrieval corpus.

Despite its rapid development and substantial potential, generative retrieval still has limitations. Generative retrieval relies on query-passage pairs for training, but the relevance judgments between queries and passages are typically incomplete. On the one hand, for a given query,  only a few passages (or even just one) are judged, while the judgments of other passages are missing; on the other hand, the judgments typically provide binary labels, which often neglect the reality that different passages typically exhibit varying levels of relevance. During generative training, generative retrieval treats the target passage (identifier) as positive and all other passages (identifiers) as equally negative. This introduces substantial noise and disrupts the training of generative retrieval, exacerbating the limitations of incomplete judgments.

Tackling the aforementioned issues is challenging, as they
are intrinsically inherent to the generative training of generative retrieval. In this work, we propose to tackle the above problem by innovating the training paradigm via knowledge distillation. We aim to employ a powerful rank (teacher) model to give multi-level judgments of quey-passage pairs for the generative retrieval (student) model. While previous studies have explored distillation for dense retrieval, its direct application to generative retrieval is not feasible due to significant differences in their training paradigms. To successfully introduce knowledge distillation into generative retrieval, two fundamental issues must be redefined and emphasized: what type of knowledge should be distilled from the teacher model, and how to effectively leverage the knowledge to benefit the student model?

\begin{figure}[t]
\centering
\includegraphics[width=1.0\linewidth]{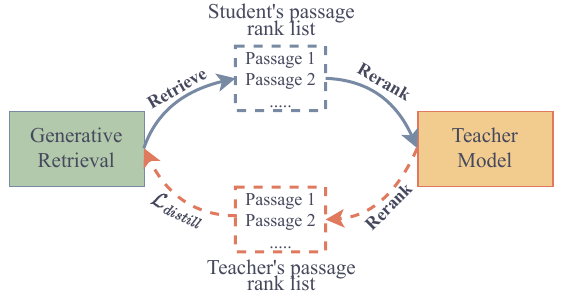}
  \vspace{-1em}
  \caption{The illustration of distillation enhanced generative retrieval (DGR) framework. Sophisticated ranking models serve as teacher models to rerank the passages, and the custom-designed distilled RankNet loss is utilized to optimize the generative retrieval model.}
  \vspace{-1em}
  \label{method}
\end{figure}

In pursuit of this goal, we introduce a Distillation-enhanced Generative Retrieval framework, dubbed DGR, as illustrated in Figure~\ref{method}. DGR characterizes knowledge as the rank order of passages provided by the teacher model and specially designs the distilled RankNet loss to efficiently incorporate this knowledge into the generative retrieval model. Specifically, we first train a typical generative retrieval model to retrieve passages. Subsequently, we utilize a sophisticated teacher model that excels in ranking passages better than generative retrieval, such as a cross-encoder, to rerank the retrieved passages. The rank order given by the teacher model contains fine-grained supervisor signals, which reflect the varying relevance degrees rather than simple binary labels. The custom-designed distilled RankNet is then employed to optimize the generative retrieval model based on the rank order provided by the teacher. During inference, we use the trained model to retrieve passages as in the typical generative retrieval. Therefore, the DGR framework only requires an additional distillation step and does not add any burden to the inference stage. We evaluate our proposed method on four widely used datasets, and the results demonstrate that DGR achieves the best performance in generative retrieval.

The key contributions are summarized:
\begin{itemize}
  \item We are the first to introduce distillation into generative retrieval, identifying a viable direction and opening doors for potential advancements in generative retrieval.
  \item We propose the DGR framework, which formulates the distilled RankNet loss to effectively incorporate knowledge from advanced rank models into generative retrieval models.
\item DGR achieves state-of-the-art performance in generative retrieval on four widely used datasets without any burden on the inference stage. DGR demonstrates exceptional robustness and generalizability with various teacher models and distillation losses.
\end{itemize}

\section{Related Work}
\subsection{Generative Retrieval}
Generative retrieval is an emerging new retrieval paradigm, which generates identifier strings of passages as the retrieval target. Instead of generating entire passages, generative retrieval uses identifiers to represent passages for reducing the amount of useless information~\cite{li2023multiview}. Previous approaches have explored different types of identifiers for different search scenarios, including titles (URLs)~\cite{de2020autoregressive, li2023generative}, numeric IDs~\cite{tay2022transformer}, substrings~\cite{bevilacqua2022autoregressive}, codebook~\cite{sun2023learning}, and multiview identifiers ~\cite{li2023multiview}. Despite the above different identifiers, generative retrieval still lags behind advanced dense retrievers. In this work, we aim to introduce knowledge distillation to further enhance generative retrieval.
\subsection{Knowledge Distillation in Text Retrieval}
Knowledge distillation is proposed by~\citeauthor{hinton2015distilling} to compress the knowledge in a powerful model into another one. In the text retrieval task, researchers also explored distilling knowledge from cross-encoder rankers into dense retrievers to enhance performance~\cite{hofstatter2020improving, hofstatter2021efficiently, qu2021rocketqa, zeng2022curriculum}, because cross-encoder rankers are more powerful in ranking passages via fine-grained interactions.  However, it is challenging to distill knowledge from cross-encoder rankers into generative retrieval, due to their different training paradigms and architectures. In this work, we propose a feasible framework to facilitate knowledge distillation in generative retrieval.
\subsection{Dense Retrieval}
Dense retrieval~\cite{lee2019latent,karpukhin2020dense} is currently the de facto implementation of text retrieval. This method benefits from the powerful representation abilities of encoder-based language models and the MIPS algorithm~\cite{shrivastava2014asymmetric}, allowing for efficient passage retrieval from a large-scale corpus. Dense retrieval has been constantly developed through knowledge distillation, hard negative sampling, and better pre-training design~\cite{chang2019pre,wang2022simlm}. Compared to dense retrieval, which relies on the dual-encoder architecture, the recently arsing generative retrieval shows promise in overcoming the missing fine-grained interaction problem through the encoder-decoder paradigm. Despite the huge potential, generative retrieval still lags behind the state-of-the-art dense retrieval method and leaves much room for investigation.

\section{Method}
In this section, we begin by giving a preliminary of a standard generative retrieval system. Following this, we outline our distillation framework as it applies to the generative retrieval system.
\subsection{Preliminary: Generative Retrieval}
\label{Preliminary: Generative Retrieval}
We present MINDER~\cite{li2023multiview}, an advanced generative retrieval system, as the base model to demonstrate the generative retrieval scheme. MINDER leverages multiview identifiers, enabling robust performance across diverse search scenarios and achieving advanced performance.

\textbf{Identifiers}. MINDER employs three types of identifiers to represent a passage: title, substring, and pseudo-query. The title generally reflects the main topic of the passage, while the substring is a randomly selected excerpt. The pseudo-query is generated based on the passage's content. These identifier types complement each other and adapt to different scenarios, contributing to MINDER's robustness across various search contexts.

\textbf{Training}. MINDER optimizes an autoregressive language model, denoted as $\textbf{AM}$, via typical sequence-to-sequence loss. The input text is the query text, and the output is the identifier of the corresponding passage that is relevant to the query. During training, the three identifiers of the samples are randomly shuffled to train the autoregressive model. For each training sample,  the objective is to minimize the sum of the negative loglikelihoods of the tokens $\{i_1,\cdots,i_l\}$ in a target identifier $I$, whose length is $l$. The generation loss is formulated as,
\begin{equation}  \label{eqn1}
   \begin{aligned}
   &\mathcal{L}_{gen} = 
   &-\sum_{j=1}^{l}\log p_{\theta}({i_j}|q;I_{<j}),
   \end{aligned}
 \end{equation}
where $I_{<j}$ denotes the partial identifier sequence $\{i_0,\cdots,i_{j-1}\}$, $q$ is the question text, and $\theta$ is the trainable parameters in the autoregessive model.

\textbf{Inference}. During the inference process, given a query text, the trained autoregressive language model $\textbf{AM}$ could generate several predicted identifiers via beam search, denoted as $\mathcal{I}$. A heuristic function is employed to transform the predicted identifiers $\mathcal{I}$ into a ranked list of passages. This function selects the predicted identifier $i_p \in \mathcal{I}_p$ for a given passage $p$ if $i_p$ appears at least once in the identifiers of that passage. The similarity score of the passage $p$ corresponding to the query $q$ is then calculated as the sum of the scores of its covered identifiers, as follows,
\begin{equation}  \label{eqn2}
   \begin{aligned}
   s_{stu} = \sum_{i_p \in \mathcal{I}_p} s_{i_p},
   \end{aligned}
 \end{equation}
where $s_{i_p}$ represents the language model score of the identifier $i_p$, and $\mathcal{I}_p$ is the set of selected identifiers that appear in the passage $p$. We just give a brief overview here, and please refer to the original paper for details.

By sorting the similarity score $s_{stu}$ of the passage, we are able to obtain a rank list of passages from the student model, denoted as $\mathcal{R}_{stu} = (p_1, ..., p_N)$, where $N$ is the number of passages in ${R}_{stu}$. 
\subsection{Distillation enhanced Generative Retrieval}
To introduce distillation into generative retrieval, we must address two key questions. First, we should determine the teacher model and what type of knowledge to be distilled into the student model (the generative retrieval model). Second, we should choose an effective distillation loss for distilling the knowledge into the student model.

\textbf{Teacher model and knowledge type}. The teacher model should be more powerful in ranking passages than the student model. We consider existing cross-encoder rankers, which enable fine-grained interaction between queries and passages, as the teacher models for generative retrieval. These rankers are trained using contrastive loss and produce logits as similarity scores. However, since generative retrieval is trained using generative loss, it cannot directly mimic the teacher's similarity scores that have different distributions.

We propose using the rank order of the passages provided by the teacher model as the distilled knowledge. Specifically, the teacher model can rerank the retrieved passages $\mathcal{R}_{stu}$ from the student model and give a new rank list $\mathcal{R}_{tea} = (p_1, ..., p_M)$, where $M$ is the number of passages in ${R}_{tea}$. The $\mathcal{R}_{tea}$ has the better rank orders attributed to the superior teacher model.  In practice, the value of $M$ is typically set to be smaller than $N$ limited by GPU memory. As a result, we sample $M$ passages from $\mathcal{R}_{stu}$ for the teacher model to rerank. We explore different sample strategies in Section~\ref{Analysis on Sample Strategies}. 

We believe the rank order of passages in $\mathcal{R}_{tea}$ is effective knowledge that could be used for distillation. Unlike the binary ``1'' and ``0'' relevance labels, the passage ranks reflect the varying degrees of relevance from the teacher model. This is evident in the different positions of negative passages in the ranking list, as opposed to having the same ``0'' label in binary labels.

\textbf{Distillation loss}. Selecting an appropriate distillation loss is the key to transferring knowledge from the teacher model to the generative retrieval model. Since we aim to distill the rank order of passages, pair-wise and list-wise rank losses seem to be natural choices. We could regard the rank order given by the teacher model as the true labels to apply pair-wise and list-wise rank losses. However, our practical findings, as detailed in Section~\ref{Analysis on Distillation Loss}, indicate that these rank losses may not effectively serve as distillation losses.

Therefore, we propose the distilled RankNet loss to better suit the DGR framework, as it can effectively leverage the rank order provided by the teacher model. For the rank list $\mathcal{R}_{tea} = (p_1, ..., p_M)$, we denote $r_i \in [1, ..., M]$ as the rank of the passage $p_i$. The distilled RankNet loss is formulated as follows,
\begin{equation}  \label{eqn3}
\left\{
   \begin{aligned}
   & \mathcal{L}_{distill} = \sum_{i=1}^M\sum_{j=1}^M \mathbbm{1}_{r_i < r_j} (s_{stu}^j - s_{stu}^i + m_{ij}), \\
   & m_{ij} = m_{base} + m_{gap}*(r_j- r_i-1),
   \end{aligned}
   \right.
 \end{equation}
where $s_{stu}^i$ represents the $i$ th passage's score provided by the student model, and $m_{ij}$ denotes the margin value determined by the rank of the two passages in the rank list ${R}_{tea}$ provided by the teacher model. $m_{base}$ is the base margin value and $m_{gap}$ is the incremental margin value. 

Similar to RankNet~\cite{burges2005learning}, our distilled RankNet loss also forms $M*(M-1)$ pairs. Differently, we adjust the RankNet loss from the following two aspects. 1) The order in which passages are ranked is determined by the teacher model rather than by the true labels. 2) The RankNet loss function only requires that the former passage have a higher score than the latter. But the distilled RankNet introduces a margin for each pair, and the margin increases as the samples are further apart in the teacher's rank list $\mathcal{R}_{tea}$. In this way, our distilled RankNet loss guarantees list-wise optimization under the guidance of the teacher model.

\textbf{Training and inference}. The final loss is defined as the interpolation of the generation loss and distillation loss: $\mathcal{L} = \alpha \mathcal{L}_{gen} + \mathcal{L}_{distill}$. 
It is noted that the gradients will be backpropagated to the generative retrieval model through the $S_{stu}$ in Equation (\ref{eqn3}). More importantly, our distillation framework optimizes the neural model and does not add any additional burden to the original inference stage. After training, the autoregressive model can be used to retrieve passages as introduced in Section~\ref{Preliminary: Generative Retrieval}.

\begin{table*}[t]
\renewcommand\arraystretch{1}
  \centering
    \scalebox{0.95}{
    \begin{tabular}{cccccccc}
    \toprule
    \multicolumn{1}{c}{\multirow{2}*{Methods}}
    &\multicolumn{3}{c}{\makecell[c]{Natural Questions}}&&\multicolumn{3}{c}{\makecell[c]{TriviaQA}}\\\cline{2-4}\cline{6-8}
         &@5&@20&@100&&@5&@20&@100\cr
    \toprule
    BM25&43.6&62.9&78.1&&67.7&77.3&83.9\cr
    DPR\cite{karpukhin2020dense}&{68.3}&{80.1}&86.1&&{72.7}&{80.2}&{84.8}\cr 
GAR\cite{mao2021generation}&59.3&73.9&85.0&&\textbf{73.1}&\textbf{80.4}&{85.7}\cr \toprule
    DSI-BART\cite{tay2022transformer}&28.3&47.3&65.5&&-&-&-\cr 
    SEAL-LM\cite{bevilacqua2022autoregressive}&40.5&60.2&73.1&&39.6&57.5&80.1\cr 
    SEAL-LM+FM\cite{bevilacqua2022autoregressive}&43.9&65.8&81.1&&38.4&56.6&80.1\cr SEAL\cite{bevilacqua2022autoregressive}&61.3&76.2&{86.3}&&66.8&77.6&84.6\cr 
    MINDER\cite{li2023multiview}&${65.8}$&${78.3}$&${86.7}$&&$68.4$&$78.1$&${84.8}$\cr
LTRGR\cite{li2023learning}&$68.8$&$80.3$&$87.1$&&$70.2$&$79.1$&$85.1$\cr
DGR&$\textbf{71.4}^{\dagger}$&$\textbf{81.4}^{\dagger}$&$\textbf{87.4}^{\dagger}$&&$72.6^{\dagger}$&$\textbf{80.4}^{\dagger}$&$\textbf{85.8}^{\dagger}$\cr
\% \textbf{improve}&3.78\%&1.37\%&0.34\%&&3.42\%&1.64\%&0.82\%\cr
    \toprule
    \end{tabular}} 
    \vspace{-1em}
    \caption{ Retrieval performance on NQ and TriviaQA. Inapplicable results are marked by “-”. The best results in each group are marked in Bold, and \textbf{$\dagger$ denotes the best result in generative retrieval}. \% improve represents the relative improvement achieved by DGR over the previously best generative retrieval method.}  \label{tab:Retrieval performance}
\end{table*}

\section{Experiments}
\subsection{Setup}
\textbf{Datasets and evaluation}. We conducted experiments on two important retrieval scenarios, QA and Web search. For the QA scenarios\, we adopted the widely used NQ~\cite{kwiatkowski2019natural} and TriviaQA~\cite{joshi2017triviaqa} datasets under the DPR~\cite{karpukhin2020dense} setting. The two datasets are based on about $21$ million Wikipedia passages. As for the Web search scenarios, we adopted the MSMARCO dataset~\cite{nguyen2016ms} and TREC Deep Learning (DL) Track 2019~\cite{craswell2020overview} and 2020~\cite{craswell2021overview}. For evaluation metrics, we used the traditional metrics for each dataset: hits@5, @20, and @100 for NQ and TriviaQA; Recall and MRR for MSMARCO; nDCG for TREC DL. As we know, we are the pioneers in evaluating generative retrieval on such comprehensive benchmarks.

\begin{table*}[t]
\renewcommand\arraystretch{1}
  \centering
    \scalebox{0.85}{
    \begin{tabular}{cccccccc}
    \toprule
    \multicolumn{1}{c}{\multirow{2}*{Methods}}&\multicolumn{1}{c}{\multirow{2}*{Model Size}}
    &\multicolumn{4}{c}{\makecell[c]{MSMARCO}}&\multicolumn{1}{c}{\makecell[c]{TREC DL 19}}&\multicolumn{1}{c}{\makecell[c]{TREC DL 20}}\\\cline{3-6}
         &&R@5&R@20&R@100&M@10&nDCG@10&nDCG@10\cr
    \toprule
    BM25&-&28.6&47.5&66.2&18.4&51.2&47.7\cr
SEAL\cite{bevilacqua2022autoregressive}&BART-Large&19.8&35.3&57.2&12.7&-&-\cr
    MINDER\cite{li2023multiview}&BART-Large&{29.5}&{53.5}&{78.7}&{18.6}&{50.6}&{39.2}\cr
    NCI\cite{wang2022neural}&T5-Base&{-}&{-}&{-}&{9.1}&{-}&{-}\cr
    DSI\cite{pradeep2023does}&T5-Base&{-}&{-}&{-}&{17.3}&{-}&{-}\cr
    DSI\cite{pradeep2023does}&T5-Large&{-}&{-}&{-}&{19.8}&{-}&{-}\cr
    LTRGR\cite{li2023learning}&BART-Large&{40.2}&{64.5}&{85.2}&{25.5}&{58.7}&{54.7}\cr
    DGR&BART-Large&\textbf{42.9}&\textbf{67.6}&\textbf{86.5}&\textbf{26.6}&\textbf{59.5}&\textbf{58.3}\cr
     \multicolumn{1}{c}{\% \textbf{improve}}&-&6.71\%&4.81\%&1.53\%&4.31\%&1.36\%&6.58\%\cr\toprule
    \end{tabular}}  
    \vspace{-0.5em}
    \caption{Retrieval performance on the MSMARCO dataset and TREC dataset. R and M denote Recall and MRR, respectively.  “-” means the result not reported in the published work. The best results in each group are marked in Bold. The baselines' results are from their respective papers. \% improve represents the relative improvement achieved by DGR over the previously best generative retrieval method.}  \label{tab:search dataset}
    \vspace{-1em}
\end{table*}

\textbf{Baselines}. We compared DGR with several generative retrieval methods, including DSI~\cite{tay2022transformer}, DSI (scaling up)~\cite{pradeep2023does}, NCI~\cite{wang2022neural}, SEAL~\cite{bevilacqua2022autoregressive}, MINDER~\cite{li2023multiview}, and LTRGR~\cite{li2023learning}. Additionally, we included the term-based method BM25, as well as DPR~\cite{karpukhin2020dense} and GAR~\cite{mao2021generation}. All baseline results were obtained from their respective papers.

\begin{table*}[t]
\renewcommand\arraystretch{1}
  \centering
    \scalebox{1.0}{
    \begin{tabular}{ccccccccc}
    \toprule
    \multicolumn{1}{c}{\multirow{2}*{Teacher Models}}&\multicolumn{1}{c}{\multirow{2}*{Architecture}}
    &\multicolumn{3}{c}{\makecell[c]{Natural Questions}}&&\multicolumn{3}{c}{\makecell[c]{TriviaQA}}\\\cline{3-5}\cline{7-9}
         &&@5&@20&@100&&@5&@20&@100\cr
    \toprule
    E5\cite{wang2022text} &Dual-encoder&70.7&81.3&87.5&&72.8&80.6&85.9\cr
    SimLM\cite{wang2023simlm} &Cross-encoder&71.4&81.4&87.4&&72.6&80.4&85.8\cr\toprule
    \end{tabular}}  
    \vspace{-1em}
    \caption{Analysis on DGR with different teacher models. Different teacher models produce the ranking list $\mathcal{R}_{tea}$.}
    \label{tab:analysis-teacher}
    \vspace{-1em}
\end{table*}

\begin{table}[t]
\renewcommand\arraystretch{1}
  \centering
    \scalebox{0.85}{
    \begin{tabular}{cccc}
    \toprule
    \multicolumn{1}{c}{\multirow{2}*{Methods}}
    &\multicolumn{3}{c}{\makecell[c]{Natural Questions}}\\\cline{2-4}
         &@5&@20&@100\cr
    \toprule
    Base model\cite{li2023multiview}&{65.8}&{78.3}&{86.7}\cr \toprule
    KL divergence&68.9&80.6&87.0\cr
    ListMLE\cite{xia2008listwise} &68.5&80.1&{87.1}\cr
    ListNet\cite{cao2007learning} &68.9&80.6&{87.0}\cr
    approxNDCG\cite{qin2010general}&68.5&79.9&87.0\cr
    RankNet\cite{burges2005learning}&68.1&80.2&87.0\cr
    LambdaLoss\cite{wang2018lambdaloss}&68.7&80.3&87.0\cr
    Distilled RankNet&$\textbf{71.4}$&$\textbf{81.4}$&$\textbf{87.4}$\cr \toprule
    \end{tabular}}  
    \vspace{-1em}
    \caption{Analysis on distillation losses. We applied KL divergence, various pair-wise and list-wise rank losses, and our proposed distilled RankNet, as the distillation loss, to the same base MINDER model.} 
    \vspace{-1em}
    \label{tab:analysis-loss}
\end{table}

\textbf{Implementation details}. To ensure a fair comparison with previous work, we utilized BART-large as our backbone. In practice, we loaded the trained autoregressive model, MINDER~\cite{li2023multiview}, and continued training it using our proposed distillation framework. We used the released SimLM (cross-encoder)\footnote{\url{https://github.com/microsoft/unilm/tree/master/simlm}} as the teacher model. We set the $M$, $N$, $m_{base}$, $m_{gap}$, $\alpha$, as 6, 200, 300, ,100, and 500, respectively.  We have trained the model several times to confirm that the improvement is not a result of random chance and present the mid one. Our experiments were conducted on four NVIDIA A5000 GPUs with 24 GB of memory.

\subsection{Retrieval Results on QA}
We summarized the results on  NQ and TriviaQA in Table~\ref{tab:Retrieval performance}. Upon analyzing the results, we had the following findings:

(1) Among the generative retrieval methods, it was observed that DSI lags behind other generative retrieval approaches due to the lack of semantic information in numeric identifiers, requiring the model to memorize the mapping from passages to their numeric IDs. Consequently, DSI faces challenges with datasets like NQ and TriviaQA, which contain over 20 million passages. MINDER surpasses SEAL by using multiview identifiers to comprehensively represent a passage. LTRGR outperforms MINDER by leveraging additional learning-to-rank training. Notably, DGR significantly outperforms all generative retrieval approaches, including LTRGR, with improvements of 2.6 and 2.4 in hits@5 on NQ and TriviaQA, respectively. DGR, based on MINDER, notably enhances its performance through the distillation framework, improving hits@5 from 65.8 to 71.4 on NQ. This underscores the effectiveness of our proposed distillation-enhanced generative retrieval framework.

(2) Most generative retrieval approaches fall behind classical sparse and dense retriever baselines such as BM25 and DPR. While LTRGR only slightly outperforms DPR on NQ, it still performs worse on TriviaQA. However, our proposed DGR has propelled generative retrieval to a new level. DGR significantly outperforms DPR on NQ and slightly surpasses DPR on TriviaQA, marking the first time that generative retrieval has surpassed DPR in the QA scenario. It is important to acknowledge that current generative retrieval approaches still exhibit a performance gap with state-of-the-art dense retrievers in the benchmarks. As a new paradigm in text retrieval, generative retrieval requires further research, and we have identified a promising direction to enhance its capabilities.

\subsection{Retrieval Results on Web Search}
We conducted experiments on the MSMARCO and TREC DL datasets to assess the performance of generative retrieval approaches on Web search. It is worth noting that evaluations of generative retrieval methods on commonly used benchmarks in this field are scarce, and we aim to advance the evaluation of generative retrieval in general search benchmarks.

Upon analyzing the results presented in Table~\ref{tab:search dataset}, several key findings have emerged. 1) Most generative retrieval methods exhibit lower performance compared to the basic BM25, and MINDER and DSI (T5-large) only marginally outperform BM25. This is attributed to the fact that web search passages are sourced from diverse web pages, often of lower quality and lacking important metadata such as titles. 2) LTRGR surpasses other generative retrieval baselines, benefiting from its learning-to-rank scheme. 3) DGR, based on the same MINDER backbone, outperforms LTRGR, which is reasonable given that DGR incorporates additional knowledge from the teacher models. 4) Current generative retrieval methods demonstrate a significant performance gap compared to the leading methods on the MSMARCO and TREC DL benchmarks, indicating the need for further efforts to address the challenges in general web search domains.

\section{Analysis}
\subsection{Analysis on Distillation Loss}
\label{Analysis on Distillation Loss}
In this work, we introduced the distilled RankNet, as outlined in Eqn.(\ref{eqn3}), to distill the ranking order from the teacher model to a generative retrieval model. Our framework also allows other distillation losses to achieve the same goal. The KL divergence is a commonly used loss in knowledge distillation, and we could also optimize the generative retrieval model using pair and list-wise rank losses, treating the given ranking order as labels. In our experiments, we applied various distillation losses, including ListMLE, ListNet, approxNDCG, RankNet, and LambdaLoss, to the base model MINDER, and the results are summarized in Table~\ref{tab:analysis-loss}.

We obtained the following findings: 1) Our proposed distillation enhanced generative retrieval framework exhibits high generalizability. Both the KL divergence and various pair-wise and list-wise losses work within our framework. We have observed that the performance of the base model could be significantly enhanced by applying our framework with different distillation losses. 2) There is a noticeable disparity between our proposed distilled RankNet loss and others. While KL divergence aims to minimize the distance between two distributions, generative retrieval and cross-encoder rankers represent distinct retrieval paradigms with differing distributions. RankNet only necessitates that the positive item has a higher similarity score than the negative one in each pair, but it lacks list-wise optimization. As a result, we have devised the distilled RankNet loss, which establishes the incremental margins for the passages in the ranking list provided by the teacher model. Our experimental results confirm the importance of our proposed distilled RankNet loss, which significantly enhances performance.
\subsection{Analysis on Teacher Models}
In our DGR framework, we selected the cross-encoder ranker, specifically SimLM (cross-encoder), as the teacher model due to its effective ranking of passages through cross-attention between queries and passages. However, there are also dense retrievers with a dual-encoder architecture that outperforms generative retrieval. This raises the question of whether these advanced dense retrievers could serve as teacher models in the DGR framework. Consequently, we have presented the retrieval performance of the DGR framework with different teacher models featuring various architectures in Table~\ref{tab:analysis-teacher}.

\begin{table}[t!]
\renewcommand\arraystretch{1}
  \centering
    \scalebox{0.95}{
    \begin{tabular}{cccc}
    \toprule
    \multicolumn{1}{c}{\multirow{2}*{Methods}}
    &\multicolumn{3}{c}{\makecell[c]{Natural Questions}}\\\cline{2-4}
         &@5&@20&@100\cr
    \toprule
    Random&69.4&81.1&87.6\cr
    Top &70.4&80.4&86.7\cr
    Top\&random &71.4&81.4&{87.4}\cr\toprule
    \end{tabular}}  
    \vspace{-0.5em}
    \caption{ Retrieval performance of different sampling strategies. } 
    \vspace{-1em}
    \label{tab:sample-strategy}
\end{table}

The results demonstrate the robustness of the DGR framework across different teacher model architectures. We observed performance improvements when applying our DGR framework to the base model using both cross-encoder and dual-encoder architectures. This is attributed to our approach of utilizing the ranking order of the teacher model as the knowledge type, which ensures the framework's resilience irrespective of the teacher model's architecture. Additionally, we noticed similar performance improvements with different teacher models. While cross-encoder rankers are expected to have more powerful ranking abilities, the observed improvement was not significantly larger. We believe this may be due to the length of $\mathcal{R}_{tea}$. As a result of GPU resource constraints, we set the length to 6, which may not fully capture the varying ranking abilities of different teachers.
\subsection{Analysis on Sampling Strategies}
\label{Analysis on Sample Strategies}
In DGR, we selected $M$ passages from the passage list $\mathcal{R}_{stu}$ to be reranked by the teacher model. We investigated the following sampling strategies:
\begin{itemize}
\item \textbf{-Random}: Randomly selecting $M$ passages from $\mathcal{R}_{stu}$.
\item \textbf{-Top}: Choosing the top-ranked $M$ passages from $\mathcal{R}_{stu}$.
\item\textbf{-Top\&Random}: Selecting one top-ranked passage and randomly selecting $M-1$ passages from $\mathcal{R}_{stu}$.
\end{itemize}

\begin{figure}[t!]
\centering
\subfigure[] {
\includegraphics[width=0.46\linewidth]{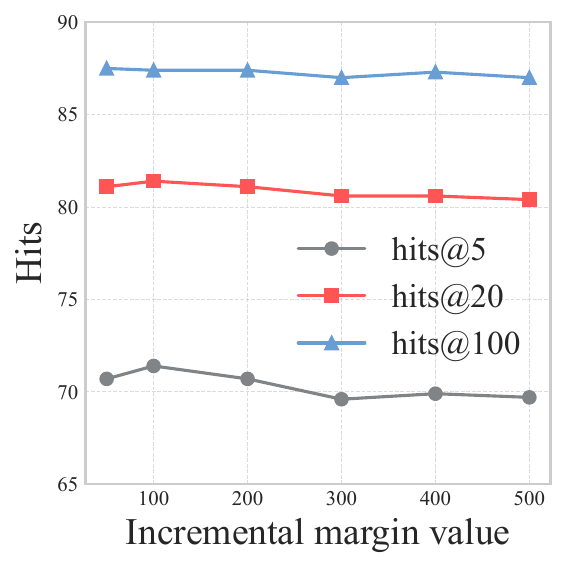}
   \label{margin_num_negative_passage_analysis:subfig1}
   }
\subfigure[] {
  \includegraphics[width=0.46\linewidth]{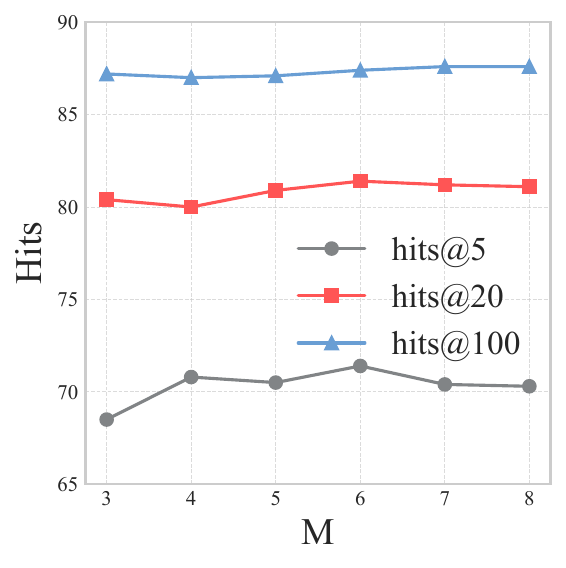}
   \label{margin_num_negative_passage_analysis:subfig2}
   }
\vspace{-1em}
\caption{Retrieval performances of DGR on the NQ test set are depicted in (a) and (b) with respect to the incremental margin values $m_{gap}$ and the number of passages, $M$, in $\mathcal{R}{tea}$.}
\vspace{-1em}
\label{margin_num_negative_passage_analysis}
\end{figure}

The results are summarized in Table~\ref{tab:sample-strategy}, and we have made several key findings. 1) When comparing the ``Random'' and ``Top'' strategies, we observed that ``Random'' performs better in hits@100, while ``Top'' achieves better hits@5 performance. This is reasonable as the ``Random'' strategy evenly selects passages from $\mathcal{R}_{stu}$, benefiting hits@100, while the ``Top'' strategy selects top-ranked passages better for hits@5. 2) The ``Top\&random'' strategy balances the two strategies and achieves better overall performance on hits@5, 20, and 100. It is worth noting that ``Top\&random'' also outperforms ``Top'' in hits@5, as only considering the top $M$ passages is too extreme, especially when $M$ is set to 6 in our experiment.
\subsection{In-depth Analysis}
\textbf{Incremental margin value $m_{gap}$}. Our proposed distilled RankNet introduces a margin for each pair, with the margin increasing as the samples are further apart in the rank list $\mathcal{R}_{tea}$, as defined in Eqn. (\ref{eqn3}). To evaluate the impact of incremental value $m_{gap}$ on retrieval performance, we manually set margin values ranging from 50 to 500 in Eq.~\ref{eqn3} and summarized the results in Figure~\ref{margin_num_negative_passage_analysis:subfig1}. Our findings indicate that the distilled RankNet with a margin of 100 performs better than the value of 50. However, as the margin value increases from 100 to 500, performance gradually declines. This suggests that a larger margin value indicates larger similarity gaps among passages in the rank list, but may not be suitable as many passages are also relevant to the query. Therefore, we looped the $m{gap}$ values to find the optimal one.

\begin{table}[t]
\renewcommand\arraystretch{1}
  \centering
    \scalebox{1.0}{
    \begin{tabular}{cccc}
    \toprule
    \multicolumn{1}{c}{\multirow{2}*{Methods}}
    &\multicolumn{3}{c}{\makecell[c]{Natural Questions}}\\\cline{2-4}
         &@5&@20&@100\cr
    \toprule
    SEAL &61.3&76.2&{86.3}\cr
    SEAL-LTR&63.7&78.1&86.4\cr
    SEAL-DGR&68.4&80.2&87.0\cr\toprule
    \end{tabular}}  
    \vspace{-1em}
    \caption{ Retrieval performance of SEAL, SEAL-LTR and SEAL-DGR on NQ. SEAL-LTR and SEAL-DGR represents applying our proposed DGR and LTRGR\cite{li2023learning} framework to the SEAL model.} 
    \vspace{-1em}
    \label{tab:SEAL-DGR}
\end{table}
\textbf{Number of passages $M$}. The number of passages in $\mathcal{R}_{tea}$ indicates how many passages the teacher model ranks. To determine the optimal length of $\mathcal{R}_{tea}$, we conducted a tuning experiment with different $M$ values, and the results are summarized in Figure~\ref{margin_num_negative_passage_analysis:subfig1}. It is observed that the retrieval performance gradually increases as $M$ increases. This is reasonable as more passages in $\mathcal{R}_{tea}$ represent more knowledge provided by the teacher model. However, it is important to note that a larger length requires more GPU memory, and we were only able to set it to 8.

\textbf{Generalization of DGR}. Our DGR builds on the generative retrieval model MINDER and continues to train it via the distillation loss. This leads to the question of whether DGR can be generalized to other generative retrieval models. To address this, we replaced MINDER with SEAL as the basic model and enhanced it using our proposed distillation framework. Additionally, we compared it with another additional training framework~\cite{li2023learning}. The results are presented in Table~\ref{tab:SEAL-DGR}. The performance of SEAL showed significant improvement after applying our DGR framework. The hits@5 metric improved from 61.3 to 68.4, confirming the effectiveness of DGR on other base generative retrieval models. Furthermore, when compared with the learning-to-rank (LTR) framework, DGR also demonstrated significant improvement, with a 4.7-point increase in hits@5. This improvement can be attributed to DGR's introduction of extra teacher models, allowing it to learn knowledge from these models.

\textbf{Inference speed}. DGR enhances base generative retrieval models, such as MINDER, through the distillation loss without impacting the inference process of the base generative retrieval model. As a result, the speed of inference remains the same as the underlying generative retrieval model. On the NQ test set, using one V100 GPU with 32GB memory, DGR, LTRGR, and MINDER took approximately 135 minutes to complete the inference process, while SEAL took only 115 minutes. It is worth noting that SEAL's speed is comparable to that of the typical dense retriever, DPR, as reported in ~\cite{bevilacqua2022autoregressive}.

\section{Conclusion and Future Work}
In this study, we proposed the distillation enhanced generative retrieval (DGR) framework, which could significantly improve generative retrieval models by distilling knowledge from advanced rank models. Importantly, DGR is highly adaptable and can effectively work with different teacher models and distillation losses. Additionally, we introduced the distilled RankNet loss, which is tailored to generative retrieval and has been evident to be more effective than existing distillation losses. We conducted extensive experiments on four widely used benchmarks to verify the effectiveness and robustness of DGR. The results verify that distillation is a promising direction for improving generative retrieval systems.

In the future, we aim to improve DGR from the following aspects. We have investigated the use of cross-encoder and dual-encoder as teacher models, and we will also explore the possibility of advanced generative retrieval models as teacher models for outdated generative retrieval models. Additionally, DGR has only begun to validate the feasibility of distillation in generative retrieval, leaving ample room for further research. Further studies, including more sampling strategies and progressive training methods, are worth exploring.

\section*{Limitations}
1) Due to limitations in GPU memory, we only assessed DGR using the maximum length of 8 for $\mathcal{R}_{tea}$. It is important to note that a longer length typically implies a greater amount of knowledge from the teacher model. Therefore, the performance of DGR could potentially be enhanced with a longer length, but we cannot verify this.
2) DGR achieves the best performance among the current generative retrieval methods, but it still falls short of the current state-of-the-art on the leaderboards. This is attributed to the model's autoregressive generation method, which generates from left to right and may not fully capture the entire content of a passage. Further techniques should be explored for generative retrieval. Fortunately, distillation appears to be a promising direction, and we have assessed its feasibility. More studies following this direction are expected.
\section*{Ethics Statement}
The datasets used in our experiment are publicly released and labeled through interaction with humans in English. In this process, user privacy is protected, and no personal information is contained in the dataset. The scientific artifacts that we used are available for research with permissive licenses. And the use of these artifacts in this paper is consistent with their intended use. Therefore, we believe that our research work meets the ethics of ACL. 
\bibliography{acl_latex}
\end{document}